\documentclass[final]{article}

\usepackage[nonatbib, final]{nips_2016}

\usepackage[utf8]{inputenc} 
\usepackage[T1]{fontenc}    
\usepackage{url}            
\usepackage{booktabs}       
\usepackage{amsfonts}       
\usepackage{nicefrac}       
\usepackage{microtype}      
\usepackage[colorinlistoftodos]{todonotes}

\usepackage{amsmath}
\usepackage{listings}
\usepackage{caption}
\usepackage{subcaption}
\usepackage{graphicx}
\usepackage{tabularx}

\title{\textsc{genesim}: genetic extraction of a single, interpretable model}

\author{
	Gilles Vandewiele, Olivier Janssens, Femke Ongenae, Filip De Turck, Sofie Van Hoecke\\
	Department of Information Technology\\
	Ghent University - imec, IDLab\\
	\texttt{gilles.vandewiele@intec.ugent.be} \\
}

\begin{document}
	
	\maketitle
	
	\begin{abstract}
		Models obtained by decision tree induction techniques excel in being interpretable. However, they can be prone to overfitting, which results in a low predictive performance. Ensemble techniques are able to achieve a higher accuracy. However, this comes at a cost of losing interpretability of the resulting model. This makes ensemble techniques impractical in applications where decision support, instead of decision making, is crucial. 
		
		To bridge this gap, we present the \textsc{genesim} algorithm that transforms an ensemble of decision trees to a single decision tree with an enhanced predictive performance by using a genetic algorithm. We compared \textsc{genesim} to prevalent decision tree induction and ensemble techniques using twelve publicly available data sets. The results show that \textsc{genesim} achieves a better predictive performance on most of these data sets than decision tree induction techniques and a predictive performance in the same order of magnitude as the ensemble techniques. Moreover, the resulting model of \textsc{genesim} has a very low complexity, making it very interpretable, in contrast to ensemble techniques.
	\end{abstract}
	
	\section{Introduction}
	
	Decision tree induction is a white-box machine learning technique that obtains an easily interpretable model after training. For each prediction from the model, an accompanying explanation can be given. Moreover, as opposed to rule extraction algorithms, the complete structure of the model is easy to analyze as it is encoded in a decision tree.
	
	In domains where the decisions that need to be made are critical, the emphasis of machine learning is on offering support and advice to the experts instead of making the decisions for them. As such, the interpretability and comprehensibility of the obtained models are of primal importance for the experts that need to base their decision on them. Therefore, a white-box approach is preferred. Examples of critical domains include the medical domain (e.g. cardiology and oncology) and the financial domain (e.g. claim management and risk assessment). 
	
	One of the disadvantages of decision trees is that they are prone to overfit \cite{slonim2002patterns}. To overcome this shortcoming, ensemble techniques have been proposed. These techniques combine the results of different classifiers \cite{Dietterich2000}, leading to an improvement in the prediction performance because of three reasons. First, when the amount of training data is small compared to the size of the hypothesis space, a learning algorithm can find many different hypotheses that correctly classify all the training data, while not performing well on unseen data. By averaging the results of the different hypotheses, the risk of choosing a wrong hypothesis can be reduced. Second, many learning algorithms can get stuck in local optima. By constructing different models from different starting points, the chance to find the global optimum is increased. Third, because of the finite size of the training data set, the optimal hypothesis can be outside of the space searched by the learning algorithm. By combining classifiers, the search space gets extended, again increasing the chance to find the optimal classifier. Nevertheless, ensemble techniques also have disadvantages. First, they take considerably longer to train and make a prediction. Second, their resulting models require more storage. The third and most important disadvantage is that the obtained model consists either out of many decision trees or only one decision tree that contains uninterpretable nodes, making it infeasible to even impossible for experts to interpret and comprehend the obtained model. To bridge the gap between decision tree induction algorithms and ensemble techniques, methods are required that can convert the ensemble into a single model. By first constructing an ensemble from the data and then applying this post-processing method, a better predictive performance can possibly be achieved than constructing a decision tree from the data directly.
	
	This post-processing technique is not only useful to increase the predictive performance while maintaining excellent interpretability. It can also be used in a big data setting where the size of the training data set is too large to construct a predictive model on a single node in a feasible amount of time. To solve this, the data set can be partitioned and a predictive model can be constructed for each of these partitions in a distributed fashion. Finally, the different models can be combined together.
	
	In this paper, we present a novel post-processing technique for ensembles, called \textsc{genesim}, which is able to convert the different models from the ensemble into a single, interpretable model. Since each of the models in the ensemble being merged will have an impact on the predictive performance of the final, combined model, a genetic approach can be applied which constructs a large ensemble and tries combining models from different subsets of this ensemble. The outline of the rest of this paper is as follows. First, in Section \ref{sec:relwork} work related to our technique and their shortcomings are presented. Then, in Section \ref{sec:genesim}, the different steps of \textsc{genesim} are depicted. In Section \ref{sec:results}, a comparison regarding predictive performance and model complexity is made between the proposed algorithm and prevalent ensemble \& decision tree induction techniques. Finally, in Section \ref{sec:conclusion}, a conclusion and possible future work are presented. 
	
	\section{Related work} \label{sec:relwork}
	
	In Van Assche et al. \cite{van2007seeing}, a technique called Interpretable Single Model (\textsc{ism}) is proposed. This technique is very similar to an induction algorithm, as it constructs a decision tree recursively top-down, by first extracting a fixed set of possible candidate tests from the trees in the ensemble. For each of these candidate tests, a split criterion is calculated by estimating the parameters using the ensemble instead of the training data. Then, the test with the optimal split criterion is chosen and the algorithm continues recursively until a pre-prune condition is met. Two shortcomings of this approach can be identified. First, information from all models, including the ones that will have a negative impact, are used to construct a final model. Second, because of the similarity with induction algorithms, it is possible to get stuck in the same local optimum as these algorithms.
	
	Deng \cite{deng2014interpreting} introduced \textsc{stel}, which converts an ensemble into an ordered rule list using the following steps. First, for each tree in the ensemble, each path from the root to a leaf is converted into a classification rule. After all rules are extracted, they are pruned and ranked to create an ordered rule list. This sorted rule set can then be used for classification by iterating over each rule and returning the target when a matching rule is found. While a good predictive performance is reported for this technique, it is much harder to grasp an ordered rule list completely than a decision tree. Therefore, when interpretability is of primal importance, the post-processing technique, that converts the ensemble of models into a single model, should result in a decision tree.
	
	A thorough survey of evolutionary algorithms for decision tree evolving can be found in \cite{Barros2012}. Evolutionary algorithms for decision trees generate an initial population of decision trees, and then crosses over the trees by replacing subtrees in one tree with subtrees of another. With a certain probability, an individual of the population can be mutated by applying operations such as replacing a subtree by a randomly generated tree, changing the information corresponding to the test in a node or swapping two subtrees in the same decision tree.
	
	\section{GENetic Extraction of a Single, Interpretable Model (\textsc{genesim})} \label{sec:genesim}
	
	While in Barros et al. \cite{Barros2012}, genetic algorithms are discussed which genetically construct decision trees from the data directly, in this paper, a genetic algorithm is applied on an ensemble of decision trees, created by using well-known induction algorithms combined with techniques including bagging and boosting. Applying a genetic approach allows to efficiently traverse the very large search space of possible model combinations. This results in an innovative approach for merging decision trees that exploits the positive properties of creating an ensemble. By exploiting multi-objective optimization, the resulting algorithm increases the accuracy \'and decreases the decision tree size at the same time, while most of the state-of-the-art succeeds in only one of the two. 
	
	Below, the different generic steps of a genetic algorithm  \cite{Sastry2005}, applied on \textsc{genesim}\footnote{\url{https://github.com/IBCNServices/GENESIM}}, are elaborated: 
	\begin{itemize}
		\item \textbf{Initialization: } to create an initial population, decision trees are generated from a training set of data using different induction algorithms, combined with ensemble techniques such as bagging and boosting. It is important that this population provides enough diversity, which allows for an extensive search space and reduces the chance of being stuck at local optima.
		\item \textbf{Evaluation: } in order to measure how `fit' a certain individual is in our population, the accuracy on a validation set is measured. In case of a tie, the model with the lowest model complexity is preferred.
		\item \textbf{Selection: } tournament selection \cite{Goldberg1989} is applied to select which individuals get combined in each iteration.
		\item \textbf{Recombination: } in order to merge two decision trees together, they are first converted to a
		set of k-dimensional hyperplanes. When all the nodes from all the trees are converted to their corresponding set of hyperplanes, the different decision spaces can be merged together by calculating their intersection using a sweep line approach discussed in \cite{Andrzejak2013}. In this approach, each hyperplane is projected on a line segment in each dimension. These line segments are then sorted, making it easy to find the intersecting line segments in a dimension. In the end, if the projected line segments of two hyperplanes intersect in each dimension, the hyperplanes intersect as well. Subsequently, their intersection can be calculated and added to the resulting decision space. This method requires $O(k * n * log(n))$ computational time, with $k$ the dimensionality of the data and $n$ the number of planes in the sets, opposed to the quadratic complexity of a naive approach which calculates the intersection of each possible pair of planes. Finally, we need to convert our merged decision space back to a decision tree. A heuristic approach is taken which identifies candidate splitting planes to create a node from, and then picks one from these candidates. To select a candidate, a metric (such as information gain) could be used, but this would introduce a bias. Therefore, a candidate is selected randomly. The candidate hyperplanes need to fulfill the constraint that they have no boundaries in all dimensions (or bounds equal to the lower and upper bound of the range of each dimension) except for one.
		
		\item \textbf{Mutation: } two possible mutations are implemented: (i) choosing a random node in the tree and replacing its threshold value by a new random number and (ii) swapping two random subtrees with eachother.
		\item \textbf{Replacement}: the population for the next iteration is created by sorting the individuals by their fitness and only selecting the first $population\_size$ individuals.
	\end{itemize} 
	
	\section{Results and evaluation} \label{sec:results}
	
	The proposed algorithm is compared, regarding the predictive performance and model complexity, to two ensemble methods (Random Forests (\textsc{rf} \cite{Breiman1984}) \& eXtreme Gradient Boosting (\textsc{xgboost} \cite{DBLP:journals/corr/ChenG16})) and four decision tree induction algorithms (\textsc{c4.5} \cite{Quinlan1993_2}, \textsc{cart} \cite{Breiman1984}, \textsc{guide} \cite{loh2009improving} and \textsc{quest} \cite{loh2008classification}). For this, twelve data sets, having very distinct properties, from the UCI Machine Learning Repository \cite{Lichman:2013} were used. An overview of the characteristics of each data set can be found in Table \ref{tbl:datasets}. 
	
	\begin{table}
		\centering
		\resizebox{\textwidth}{!}{%
			\begin{tabular}{ccccc||ccccc}
				\hline \textbf{name} & \textbf{\#samples} & \textbf{\#cont} & \textbf{\#disc} & \textbf{class\_dist} & \textbf{name} & \textbf{\#samples} & \textbf{\#cont} & \textbf{\#disc} & \textbf{class\_dist} \\ \hline
				iris &150 &4 &0 &33.3 - 33.3 - 33.3&austra &690 &5 &9 &55.5 - 44.5  \\
				cars &1727 &0 &6 &70.0 - 22.2 - 4.0 - 3.8  & ecoli &326 &5 &2 &43.6 - 23.6 - 16.0 - 10.7 - 6.1  \\
				glass &213 &9 &0 &32.4 - 35.7 - 8.0 - 6.1 - 4.2 - 13.6  & heart &269 &5 &8 &55.8 - 44.2  \\
				led7 &2563 &0 &7 &13 - 13 - 12 - 11 - 13 - 13 - 13 - 12 & lymph &142 &0 &18 &57.0 - 43.0  \\
				pima &768 &7 &1 &65.1 - 34.9  & vehicle &846 &14 &4 &25.1 - 25.7 - 25.8 - 23.5  \\
				wine &177 &13 &0 &32.8 - 40.1 - 27.1  & breast &698 &0 &9 &65.5 - 34.5  \\ \hline
			\end{tabular}}
			\caption{Table with the characteristics for each data set}
			\label{tbl:datasets}
	\end{table}
	
\begin{figure}[h]
		\begin{subfigure}{.45\textwidth}
			\centering
			\includegraphics[width=\textwidth]{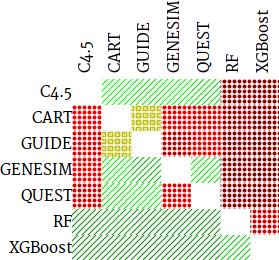}
			\subcaption{WTL matrix for the accuracies}
			\label{fig:wtl_acc}
		\end{subfigure}
		\begin{subfigure}{.45\textwidth}
			\centering
			\includegraphics[width=\textwidth]{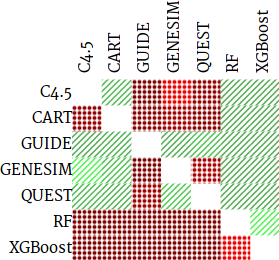}
			\subcaption{WTL matrix for the average number of nodes (or trees) in the resulting model}
			\label{fig:wtl_compl}
		\end{subfigure}
		\caption{Win-Tie-Loss matrices for the different algorithms for accuracies and model complexities}
		\label{fig:wtl}	
	\end{figure}

	The hyper-parameters of each of the tree induction and ensemble techniques were tuned using grid search when the number of parameters was lower than four, else bayesian optimization was used. Unfortunately, because of a rather high complexity of \textsc{genesim}, hyper-parameter optimization could not be applied. The ensemble that was transformed into a single model by \textsc{genesim} was constructed using different induction algorithms (\textsc{c4.5}, \textsc{cart}, \textsc{quest} and \textsc{guide}) combined with bagging and boosting. We applied 3-fold cross validation 10 times on each of the data sets and stored the mean accuracy and model complexity for the 3 folds. The mean accuracy and mean model complexity (and their corresponding standard deviations) over these 10 measurements can be found in Table \ref{tbl:acc} and Table \ref{tbl:compl}. Bootstrap statistical significance testing was applied to construct a Win-Tie-Loss matrix, which can be seen in Figure \ref{fig:wtl}. Algorithm A wins over B for a certain data set when the mean accuracy is higher than B on that data set and the $\rho$-value for the statistical test is lower than 0.05. When an algorithm has more wins than losses compared to another algorithm, the cell is colored green (and hatched with stripes). Else, the cell is colored red (and hatched with dots). The darker the green, the more wins the algorithm has over the other. Similarly, the darker the red, the more losses an algorithm has over the other.
	
	A few things can be deduced from these matrices. First, we can clearly see that the ensemble techniques \textsc{rf} and \textsc{xgboost} have a superior accuracy compared to all other algorithms on these data sets, and that \textsc{xgboost} performs better than \textsc{rf}. While the accuracy is indeed better, the increase can be of a rather moderate size while the resulting model is completely uninterpretable. Second, in terms of accuracy, the proposed \textsc{genesim} is better than all decision tree induction algorithms, except \textsc{c4.5}. Although, \textsc{genesim} is very competitive to it (winning on two data sets while losing on three) and \textsc{c4.5} could be better due to the fact that no hyper-parameter optimization was applied to \textsc{genesim}. For each data set, the same hyperparameters were used (such as a limited amount of iterations and using 50\% of the training data as validation data).  Third, \textsc{genesim} produces very interpretable models with a very low model complexity (expressed here as the number of nodes in the tree). The average number of nodes in the resulting tree is lower than in \textsc{cart} and \textsc{c4.5}, but higher than \textsc{quest} and \textsc{guide}. But the predictive performance of the two last-mentioned algorithms is much lower than \textsc{genesim}.
	
	\section{Conclusion} \label{sec:conclusion}
	In this paper, a technique called \textsc{genesim} is proposed. While exploiting the positive properties of constructing ensembles, it results in a single, interpretable model which is ideally suited to support experts in critical domains. Results show that in most cases, an increased predictive performance can be achieved, while having a model complexity similar to the complexity of trees produced by induction algorithms. Results of \textsc{genesim} can still be improved by reducing the computational complexity of our algorithm, allowing hyper-parameter optimization and our technique to run for more iterations in a feasible amount of time. Moreover, in the future, an implementation of similar techniques, such as \textsc{ism}, to allow a comparison with \textsc{genesim} can be performed. 
	
	\begin{table}[h]
		\centering
		\resizebox{\textwidth}{!}{%
			\begin{tabularx}{\textwidth}{|l|X|X|X|X|X|X|X|X|}
				\hline & \multicolumn{1}{|c|}{{\tiny \textbf{XGB}}} & \multicolumn{1}{|c|}{{\tiny \textbf{CART}}} & \multicolumn{1}{|c|}{{\tiny \textbf{QUEST}}} & \multicolumn{1}{|c|}{{\tiny \textbf{GENESIM}}} & \multicolumn{1}{|c|}{{\tiny \textbf{RF}}} & \multicolumn{1}{|c|}{{\tiny \textbf{ISM}}} & \multicolumn{1}{|c|}{{\tiny \textbf{C4.5}}} & \multicolumn{1}{|c|}{{\tiny \textbf{GUIDE}}} \\ \hline 
				\textbf{heart}  & 0.8257\newline$\pm$0.01$\sigma$ & 0.7441\newline$\pm$0.02$\sigma$ & 0.7585\newline$\pm$0.02$\sigma$ & 0.7982\newline$\pm$0.02$\sigma$ & 0.8129\newline$\pm$0.01$\sigma$ & 0.8024\newline$\pm$0.02$\sigma$ & 0.7877\newline$\pm$0.03$\sigma$ & 0.7829\newline$\pm$0.02$\sigma$ \\ \hline 
				\textbf{led7}  & 0.8018\newline$\pm$0.0$\sigma$ & 0.7997\newline$\pm$0.0$\sigma$ & 0.7986\newline$\pm$0.0$\sigma$ & 0.7926\newline$\pm$0.0$\sigma$ & 0.8027\newline$\pm$0.0$\sigma$ & 0.7996\newline$\pm$0.0$\sigma$ & 0.8012\newline$\pm$0.0$\sigma$ & 0.761\newline$\pm$0.01$\sigma$ \\ \hline 
				\textbf{iris}  & 0.9505\newline$\pm$0.01$\sigma$ & 0.9504\newline$\pm$0.01$\sigma$ & 0.9562\newline$\pm$0.0$\sigma$ & 0.9463\newline$\pm$0.01$\sigma$ & 0.95\newline$\pm$0.01$\sigma$ & 0.9519\newline$\pm$0.01$\sigma$ & 0.9395\newline$\pm$0.01$\sigma$ & 0.9467\newline$\pm$0.01$\sigma$ \\ \hline 
				\textbf{cars}  & 0.9842\newline$\pm$0.0$\sigma$ & 0.9749\newline$\pm$0.0$\sigma$ & 0.9411\newline$\pm$0.01$\sigma$ & 0.9543\newline$\pm$0.01$\sigma$ & 0.9701\newline$\pm$0.01$\sigma$ & 0.9685\newline$\pm$0.0$\sigma$ & 0.966\newline$\pm$0.0$\sigma$ & 0.9426\newline$\pm$0.01$\sigma$ \\ \hline 
				\textbf{ecoli}  & 0.8651\newline$\pm$0.01$\sigma$ & 0.8196\newline$\pm$0.02$\sigma$ & 0.8195\newline$\pm$0.01$\sigma$ & 0.8325\newline$\pm$0.02$\sigma$ & 0.8486\newline$\pm$0.01$\sigma$ & 0.7507\newline$\pm$0.04$\sigma$ & 0.817\newline$\pm$0.03$\sigma$ & 0.8319\newline$\pm$0.01$\sigma$ \\ \hline 
				\textbf{glass}  & 0.7494\newline$\pm$0.02$\sigma$ & 0.6667\newline$\pm$0.03$\sigma$ & 0.649\newline$\pm$0.03$\sigma$ & 0.6696\newline$\pm$0.03$\sigma$ & 0.7526\newline$\pm$0.03$\sigma$ & 0.6489\newline$\pm$0.03$\sigma$ & 0.6763\newline$\pm$0.03$\sigma$ & 0.6557\newline$\pm$0.02$\sigma$ \\ \hline 
				\textbf{austra}  & 0.8686\newline$\pm$0.01$\sigma$ & 0.8506\newline$\pm$0.01$\sigma$ & 0.8547\newline$\pm$0.01$\sigma$ & 0.8553\newline$\pm$0.01$\sigma$ & 0.8663\newline$\pm$0.01$\sigma$ & 0.8557\newline$\pm$0.01$\sigma$ & 0.8528\newline$\pm$0.01$\sigma$ & 0.8582\newline$\pm$0.01$\sigma$ \\ \hline 
				\textbf{vehicle}  & 0.7606\newline$\pm$0.01$\sigma$ & 0.6988\newline$\pm$0.01$\sigma$ & 0.6986\newline$\pm$0.01$\sigma$ & 0.6834\newline$\pm$0.01$\sigma$ & 0.7383\newline$\pm$0.01$\sigma$ & 0.6672\newline$\pm$0.01$\sigma$ & 0.7115\newline$\pm$0.01$\sigma$ & 0.6821\newline$\pm$0.01$\sigma$ \\ \hline 
				\textbf{breast}  & 0.9591\newline$\pm$0.0$\sigma$ & 0.94\newline$\pm$0.01$\sigma$ & 0.947\newline$\pm$0.01$\sigma$ & 0.9496\newline$\pm$0.01$\sigma$ & 0.958\newline$\pm$0.01$\sigma$ & 0.9466\newline$\pm$0.0$\sigma$ & 0.9443\newline$\pm$0.0$\sigma$ & 0.937\newline$\pm$0.01$\sigma$ \\ \hline 
				\textbf{lymph}  & 0.8354\newline$\pm$0.02$\sigma$ & 0.7686\newline$\pm$0.02$\sigma$ & 0.7907\newline$\pm$0.03$\sigma$ & 0.7866\newline$\pm$0.02$\sigma$ & 0.817\newline$\pm$0.02$\sigma$ & 0.7822\newline$\pm$0.03$\sigma$ & 0.7839\newline$\pm$0.03$\sigma$ & 0.7659\newline$\pm$0.04$\sigma$ \\ \hline 
				\textbf{pima}  & 0.7543\newline$\pm$0.01$\sigma$ & 0.7174\newline$\pm$0.02$\sigma$ & 0.7385\newline$\pm$0.01$\sigma$ & 0.7266\newline$\pm$0.01$\sigma$ & 0.7626\newline$\pm$0.01$\sigma$ & 0.7346\newline$\pm$0.01$\sigma$ & 0.7348\newline$\pm$0.01$\sigma$ & 0.7285\newline$\pm$0.02$\sigma$ \\ \hline 
				\textbf{wine}  & 0.9709\newline$\pm$0.01$\sigma$ & 0.9072\newline$\pm$0.01$\sigma$ & 0.9055\newline$\pm$0.03$\sigma$ & 0.9128\newline$\pm$0.03$\sigma$ & 0.9603\newline$\pm$0.01$\sigma$ & 0.8838\newline$\pm$0.01$\sigma$ & 0.9217\newline$\pm$0.01$\sigma$ & 0.8828\newline$\pm$0.03$\sigma$ \\ \hline 
			\end{tabularx}}
			\caption{Mean accuracies for the different data sets and algorithms using 10 measurements}
			\label{tbl:acc} \vspace{2em}
			\resizebox{\textwidth}{!}{%
			\begin{tabularx}{\textwidth}{|l|X|X|X|X|X|X|X|X|}
				\hline & \multicolumn{1}{|c|}{{\tiny \textbf{XGB(*)}}} & \multicolumn{1}{|c|}{{\tiny \textbf{CART}}} & \multicolumn{1}{|c|}{{\tiny \textbf{QUEST}}} & \multicolumn{1}{|c|}{{\tiny \textbf{GENESIM}}} & \multicolumn{1}{|c|}{{\tiny \textbf{RF(*)}}} & \multicolumn{1}{|c|}{{\tiny \textbf{ISM}}} & \multicolumn{1}{|c|}{{\tiny \textbf{C4.5}}} & \multicolumn{1}{|c|}{{\tiny \textbf{GUIDE}}} \\ \hline 
				\textbf{heart}  & 408.4815\newline$\pm$188.2$\sigma$ & 35.8148\newline$\pm$12.54$\sigma$ & 9.1852\newline$\pm$2.97$\sigma$ & 17.4444\newline$\pm$4.84$\sigma$ & 448.6113\newline$\pm$154.6$\sigma$ & 35.8889\newline$\pm$10.71$\sigma$ & 23.5556\newline$\pm$6.62$\sigma$ & 9.1481\newline$\pm$2.28$\sigma$ \\ \hline 
				\textbf{led7}  & 459.9792\newline$\pm$152.2$\sigma$ & 201.9583\newline$\pm$1.2$\sigma$ & 57.625\newline$\pm$4.91$\sigma$ & 92.0417\newline$\pm$17.08$\sigma$ & 516.25\newline$\pm$155.4$\sigma$ & 111.2917\newline$\pm$15.45$\sigma$ & 58.9583\newline$\pm$2.09$\sigma$ & 32.9167\newline$\pm$2.55$\sigma$ \\ \hline 
				\textbf{iris}  & 544.5238\newline$\pm$144.6$\sigma$ & 12.2857\newline$\pm$1.34$\sigma$ & 5.8571\newline$\pm$0.59$\sigma$ & 5.9048\newline$\pm$0.65$\sigma$ & 453.2381\newline$\pm$204.4$\sigma$ & 10.5714\newline$\pm$1.91$\sigma$ & 7.3809\newline$\pm$1.06$\sigma$ & 5.3333\newline$\pm$0.55$\sigma$ \\ \hline 
				\textbf{cars}  & 631.2821\newline$\pm$123.7$\sigma$ & 140.1282\newline$\pm$2.66$\sigma$ & 45.6667\newline$\pm$4.7$\sigma$ & 103.1539\newline$\pm$14.42$\sigma$ & 438.4615\newline$\pm$178.3$\sigma$ & 131.4102\newline$\pm$9.62$\sigma$ & 98.4359\newline$\pm$4.6$\sigma$ & 43.6154\newline$\pm$5.07$\sigma$ \\ \hline 
				\textbf{ecoli}  & 487.5625\newline$\pm$202.9$\sigma$ & 35.6667\newline$\pm$11.77$\sigma$ & 14.5833\newline$\pm$3.48$\sigma$ & 19.0833\newline$\pm$4.27$\sigma$ & 447.0623\newline$\pm$147.7$\sigma$ & 60.125\newline$\pm$16.06$\sigma$ & 19.25\newline$\pm$2.84$\sigma$ & 10.0833\newline$\pm$1.43$\sigma$ \\ \hline 
				\textbf{glass}  & 530.7017\newline$\pm$179.2$\sigma$ & 57.8421\newline$\pm$11.27$\sigma$ & 22.4035\newline$\pm$5.66$\sigma$ & 29.6667\newline$\pm$5.75$\sigma$ & 486.9825\newline$\pm$160$\sigma$ & 80.3684\newline$\pm$24.1$\sigma$ & 36.2982\newline$\pm$3.09$\sigma$ & 16.1579\newline$\pm$2.47$\sigma$ \\ \hline 
				\textbf{austra}  & 433.0392\newline$\pm$72.7$\sigma$ & 7.7451\newline$\pm$6.19$\sigma$ & 7.902\newline$\pm$3.23$\sigma$ & 23.7843\newline$\pm$7.37$\sigma$ & 396.3333\newline$\pm$181.5$\sigma$ & 38.8824\newline$\pm$15.73$\sigma$ & 26.7255\newline$\pm$6.82$\sigma$ & 8.2941\newline$\pm$3.12$\sigma$ \\ \hline 
				\textbf{vehicle}  & 465.6667\newline$\pm$119.4$\sigma$ & 177.1111\newline$\pm$22.26$\sigma$ & 81.7778\newline$\pm$14.85$\sigma$ & 83.2222\newline$\pm$9.68$\sigma$ & 485.2778\newline$\pm$146.8$\sigma$ & 345.5556\newline$\pm$45.92$\sigma$ & 92.4444\newline$\pm$12.43$\sigma$ & 33.2222\newline$\pm$8.71$\sigma$ \\ \hline 
				\textbf{breast}  & 563.3333\newline$\pm$170.6$\sigma$ & 30.619\newline$\pm$7.89$\sigma$ & 12.619\newline$\pm$3.73$\sigma$ & 18.5238\newline$\pm$3.49$\sigma$ & 395.5714\newline$\pm$161.4$\sigma$ & 43.7619\newline$\pm$13.31$\sigma$ & 19.4762\newline$\pm$2.38$\sigma$ & 10.4286\newline$\pm$1.65$\sigma$ \\ \hline 
				\textbf{lymph}  & 608.4375\newline$\pm$140.5$\sigma$ & 32.0417\newline$\pm$5.75$\sigma$ & 13.5417\newline$\pm$3.14$\sigma$ & 14.8333\newline$\pm$4.0$\sigma$ & 497.9375\newline$\pm$162.3$\sigma$ & 30.9583\newline$\pm$6.6$\sigma$ & 16.9583\newline$\pm$2.44$\sigma$ & 8.875\newline$\pm$2.81$\sigma$ \\ \hline 
				\textbf{pima}  & 180.0556\newline$\pm$85.5$\sigma$ & 52.4445\newline$\pm$19.8$\sigma$ & 12.0\newline$\pm$4.32$\sigma$ & 45.2222\newline$\pm$8.53$\sigma$ & 434.8334\newline$\pm$68.04$\sigma$ & 101.6667\newline$\pm$18.5$\sigma$ & 26.0\newline$\pm$5.12$\sigma$ & 8.1111\newline$\pm$2.36$\sigma$ \\ \hline 
				\textbf{wine}  & 487.0948\newline$\pm$176.9$\sigma$ & 13.4762\newline$\pm$1.58$\sigma$ & 9.1905\newline$\pm$1.66$\sigma$ & 8.0476\newline$\pm$0.93$\sigma$ & 409.2381\newline$\pm$116.1$\sigma$ & 33.3809\newline$\pm$3.04$\sigma$ & 9.381\newline$\pm$0.33$\sigma$ & 6.8095\newline$\pm$0.77$\sigma$ \\ \hline 
			\end{tabularx}}
			\caption{Mean model complexities, expressed as either number of nodes in the resulting decision tree or number of decision trees in the ensemble (*), for the different data sets and algorithms using 10 measurements}
			\label{tbl:compl}
		\end{table}
	
	\clearpage
	\bibliographystyle{unsrt}
	\bibliography{GENESIM_NIPS}

\begin{thebibliography}{10}

\bibitem{slonim2002patterns}
Donna~K Slonim.
\newblock From patterns to pathways: gene expression data analysis comes of
  age.
\newblock {\em Nature genetics}, 32:502--508, 2002.

\bibitem{Dietterich2000}
Thomas~G. Dietterich.
\newblock {\em Multiple Classifier Systems: First International Workshop},
  chapter Ensemble Methods in Machine Learning, pages 1--15.
\newblock Springer Berlin Heidelberg, 2000.

\bibitem{van2007seeing}
Anneleen Van~Assche and Hendrik Blockeel.
\newblock Seeing the forest through the trees.
\newblock In {\em International Conference on Inductive Logic Programming},
  pages 269--279. Springer, 2007.

\bibitem{deng2014interpreting}
Houtao Deng.
\newblock Interpreting tree ensembles with intrees.
\newblock {\em arXiv preprint arXiv:1408.5456}, 2014.

\bibitem{Barros2012}
Rodrigo~Coelho Barros, Marcio~Porto Basgalupp, Andre C P L~F {De Carvalho}, and
  Alex~a. Freitas.
\newblock {A survey of evolutionary algorithms for decision-tree induction}.
\newblock {\em IEEE Transactions on Systems, Man and Cybernetics Part C:
  Applications and Reviews}, 42(3):291--312, 2012.

\bibitem{Sastry2005}
Kumara Sastry, David Goldberg, and Graham Kendall.
\newblock {Search Methodologies}.
\newblock {\em Compute}, pages 97--125, 2005.

\bibitem{Goldberg1989}
David~E Goldberg, Bradley Korb, and Kalyanmoy Deb.
\newblock {Messy Genetic Algorithms : Motivation , Analysis , and First
  Results}.
\newblock {\em Engineering}, 3:493--530, 1989.

\bibitem{Andrzejak2013}
Artur Andrzejak, Felix Langner, and Silvestre Zabala.
\newblock {Interpretable models from distributed data via merging of decision
  trees}.
\newblock {\em Proceedings of the 2013 IEEE Symposium on Computational
  Intelligence and Data Mining, CIDM 2013 - 2013 IEEE Symposium Series on
  Computational Intelligence, SSCI 2013}, pages 1--9, 2013.

\bibitem{Breiman1984}
Leo Breiman, Jerome Friedman, Charles~J. Stone, and R.A. Olshen.
\newblock {\em {Classification and Regression Trees}}.
\newblock Chapman and Hall/CRC, 1984.

\bibitem{DBLP:journals/corr/ChenG16}
Tianqi Chen and Carlos Guestrin.
\newblock Xgboost: {A} scalable tree boosting system.
\newblock {\em CoRR}, abs/1603.02754, 2016.

\bibitem{Quinlan1993_2}
J.~Ross Quinlan.
\newblock {\em {C4.5: programs for machine learning}}.
\newblock Morgan Kaufmann Publishers Inc., 1993.

\bibitem{loh2009improving}
Wei-Yin Loh.
\newblock Improving the precision of classification trees.
\newblock {\em The Annals of Applied Statistics}, pages 1710--1737, 2009.

\bibitem{loh2008classification}
Wei-Yin Loh.
\newblock Classification and regression tree methods.
\newblock {\em Encyclopedia of statistics in quality and reliability}, 2008.

\bibitem{Lichman:2013}
M.~Lichman.
\newblock {UCI} machine learning repository, 2013.

\end{thebibliography}
\end{document}